\documentclass[10pt,twocolumn,letterpaper]{article}

\usepackage{cvpr}              

\pdfoutput=1
\usepackage[dvipsnames]{xcolor}

\usepackage{upgreek}
\usepackage{colortbl}
\usepackage{multirow}

\newcommand{\smat}{supplementary material}

\usepackage{ifthen}
\newboolean{show_main}
\newboolean{show_supp}

\usepackage{soul}
\usepackage{comment}

\definecolor{cvprblue}{rgb}{0.21,0.49,0.74}
\usepackage[pagebackref,breaklinks,colorlinks,citecolor=cvprblue]{hyperref}

\usepackage{graphicx}
\graphicspath{ {./images/} }

\def\paperID{418} 
\def\confName{3DV\xspace}
\def\confYear{2025\xspace}

\title{Object Agnostic 3D Lifting in Space and Time}

\author{
Christopher Fusco$^1$ \and
Shin-Fang Ch'ng$^1$ \and
Mosam Dabhi$^2$ \and
Simon Lucey$^1$ \and
\\
$^1$The University of Adelaide \hspace{0.7cm}
$^2$Carnegie Mellon University \\
}

\setboolean{show_main}{true}
\setboolean{show_supp}{true}

\begin{document}
\ifthenelse{\boolean{show_main}}{
\twocolumn[{%
\renewcommand\twocolumn[1][]{#1}%
\maketitle
\vspace{-0.6cm}
\begin{center}
    \centering
    \captionsetup{type=figure}
    \includegraphics[width=\linewidth]{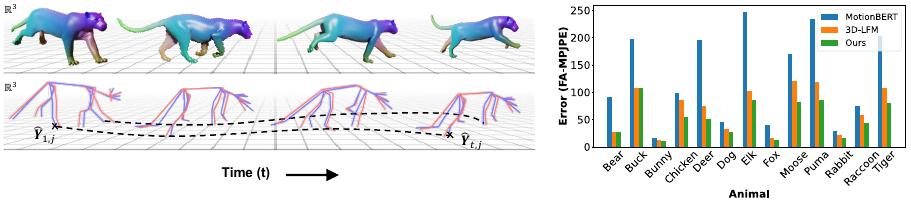}
    
    \captionof{figure}{
    \textbf{Left:} Bottom row shows the 3D skeletons of a puma animal in motion. The blue lines represent our model’s predictions, closely tracking the red ground-truth lines, demonstrating our model's ability to generate smooth and precise motion over time. The dashed line highlights the trajectory of a specific joint $\hat{\mathbf{Y}}_{t,j}$, emphasizing the temporal consistency and accuracy of our approach. \textbf{Right:} Quantitative FA-MPJPE comparison across 13 animal categories, where our method consistently outperforms competing models.}
    \label{fig:main}
\end{center}%
}]
\maketitle
\begin{abstract}

We present a spatio-temporal perspective on category-agnostic 3D lifting of 2D keypoints over a temporal sequence. Our approach differs from existing state-of-the-art methods that are either: (i) object-agnostic, but can only operate on individual frames, or (ii) can model space-time dependencies, but are only designed to work with a single object category. Our approach is grounded in two core principles. First, general information about similar objects can be leveraged to achieve better performance when there is little object-specific training data. Second, a temporally-proximate context window is advantageous for achieving consistency throughout a sequence. These two principles allow us to outperform current state-of-the-art methods on per-frame and per-sequence metrics for a variety of animal categories. Lastly, we release a new synthetic dataset containing 3D skeletons and motion sequences for a variety of animal categories. 

\end{abstract}    
\section{Introduction}
\label{sec:intro}

Reconstructing 3D deforming objects from 2D landmarks obtained by a single camera is a long-standing challenge in computer vision. Traditional non-rigid structure-from-motion (NRSfM) approaches relied on clever but straightforward factorisation methods that are sensitive to noise and occlusions \cite{tomasiShapeMotionImage1992, breglerRecoveringNonrigid3D2000}. In some cases, ambiguities can be resolved with multiple camera views, but come at the cost of expensive equipment and limited practicability to natural scenes. Recent learning-based methods are capable of robustly recovering 3D object structure from a single camera in the presence of noise and occlusions. Coupled with an abundance of publicly available 3D human pose data, human-specific lifting models like MotionBERT \cite{zhuMotionBERTUnifiedPerspective2023} have become increasingly capable. However, their reliance on human-specific information and vast amounts of training data make it problematic if they are to be used for other objects. In particular, animals pose a significant challenge due to the limited amount of publicly available 3D animal data.

This has motivated recent developments around object-agnostic lifting, where a single model is capable of lifting various object categories without category-specific fine-tuning. Most notably, 3D-LFM \cite{dabhi3DLFMLiftingFoundation2024} achieved state-of-the-art performance on a combined dataset of various object categories. The permutation equivariant property of transformers and additional skeletal information were leveraged to robustly handle category imbalances and within-category object rig/skeleton variations. 
However, its inability to utilize temporal information results in poor performance when applied to a sequence of 2D poses obtained from a video. We observe frequent jitter and poor recovery of occluded points, particularly in dynamic sequences.

Our work tackles these challenges by introducing the first object-agnostic 3D lifting framework that is both data-efficient and temporally aware. Our approach leverages the power of transformers with a strategic inductive bias that focuses attention on temporally proximate frames, enabling it to effectively capture motion dynamics. Our approach improves the accuracy of 3D reconstructions across various object categories, particularly in challenging scenarios that contain occlusions, fast movement, limited data, and previously unseen categories. Furthermore, we address the lack of publicly available datasets for lifting diverse animal skeletons and motion sequences by creating a new synthetic dataset. Our dataset, \textbf{AnimalSyn3D}, includes 4D labels for 13 animal categories, encompassing 678 animation sequences with temporal consistency, designed to enable further research in class-agnostic lifting.

The contributions of this paper are:

\begin{itemize}
    \item We propose a class-agnostic lifting model with a strategic inductive bias directly embedded in the architecture. We validate the state-of-the-art 3D lifting performance of our approach across challenging scenarios involving noise, occlusions, and unseen objects.
    \item We contribute a new synthetic dataset containing 4D skeletons for a variety of animals with animated behavior sequences, where temporal consistency is prioritized through a non-linear refinement procedure.
\end{itemize}

We empirically validate the effectiveness of our approach on our synthetic dataset. We achieve state-of-the-art results with existing metrics and provide an additional metric for a more complete analysis.
\section{Related Works}
\label{sec:related_works}

\subsection{3D Pose Estimation}
Obtaining the 3D pose of an object from a single monocular camera generally follows one of two paths. The first directly predicts the 3D pose from RGB images \cite{brau3DHumanPose2016,park3DHumanPose2016,moonCameraDistanceawareTopdown2019,pavlakosOrdinalDepthSupervision2018}, often struggling to generalize to distribution shifts such as lighting and background information. Alternatively, two-stage methods divide the task between two specialized models \cite{hossainExploitingTemporalInformation2018a,changPoseLifterAbsolute3D2020,liuAttentionMechanismExploits2020,chenAnatomyaware3DHuman2021}: a pose detector first extracts a 2D pose which is then lifted into 3D by a separate model. Our work aligns with a two-stage approach, particularly focusing on improving the robustness and generalization of the 3D lifting stage.

\subsection{Object-Specific Lifting}
Traditional NRSfM algorithms have been effective in modeling simple and targeted objects, such as human bodies and hands \cite{chenJointwise2D3D2023,ge3DHandShape2019}. These methods largely rely on the availability of 2D keypoints and specific 3D supervision for the object in question. However, recent deep learning approaches have demonstrated superior performance in handling the complexity of various object rigs \cite{novotnyC3DPOCanonical3D2019a,jiUnsupervised3DPose2023a,wangPAULProcrusteanAutoencoder2021a,kongDeepNonRigidStructure2019a}. Despite these advancements, they still require the 2D keypoints to have consistent semantic correspondence across all instances of the object, where a specific landmark, such as an elbow, must have the same semantic meaning across different poses.

This limitation persists even in state-of-the-art deep lifting models like MotionBERT \cite{zhuMotionBERTUnifiedPerspective2023} and others \cite{changPoseLifterAbsolute3D2020,chenAnatomyaware3DHuman2021}, which are tailored specifically for the human body. The specialized nature of these models and their dependence on large datasets make them unsuitable for objects with limited available data, such as animals. Existing animal-specific lifting models suffer from similar issues, often being restricted to a single animal category and demonstrating poor generalization due to data scarcity \cite{karashchukAniposeToolkitRobust2021,mathisMarkerlessTrackingUserdefined2018,gosztolaiLiftPose3DDeepLearningbased2020}.

\subsection{Object-Agnostic Lifting}
The paradigm of object-agnostic lifting has recently been pioneered by 3D-LFM \cite{dabhi3DLFMLiftingFoundation2024}, which can handle a wide range of object categories by leveraging large-scale data to enhance performance for underrepresented or unseen objects. However, unlike object-specific models such as Motionbert~\cite{zhuMotionBERTUnifiedPerspective2023}, 3D-LFM does not incorporate temporal information, which is important for accurate 3D reconstruction of sequences. Our framework builds upon this insight, integrating the strengths of both object-agnostic lifting and temporal modeling.

\subsection{Animal Datasets}
To effectively benchmark category-agnostic lifting over videos of animals, a diverse set of animal categories with accurately labeled 3D skeletons is required. Recent animal datasets have made strides in this area but are often limited to single atemporal images \cite{xuAnimal3DComprehensiveDataset2024,yaoOpenMonkeyChallengeDatasetBenchmark2023}, with few publicly available datasets offering 3D poses of animals in video sequences \cite{liPosesEquineResearch2024,chengMarmoPoseDeepLearningBased2024}. Moreover, methods to collect data for a broad set of animals are typically impractical, expensive, or yield noisy results.

Inspired by the tracking community, we turn to synthetic data \cite{doerschTAPVidBenchmarkTracking2023a,karaevDynamicStereoConsistentDynamic2023a,zhengPointOdysseyLargeScaleSynthetic2023a}. Existing synthetic datasets are restricted to a single animal, such as pigs \cite{anThreedimensionalSurfaceMotion2023} or ants \cite{plumReplicAntPipelineGenerating2023}, and generally contain only simplistic motion sequences. The DeformableThings4D \cite{li4DCompleteNonRigidMotion2021} dataset offers more complex and diverse motion sequences animated by artists, but it was created for dense mesh recovery and does not include 3D skeletons. We build upon these models and animations to create a new dataset specifically designed for the task of class-agnostic 3D lifting.

\section{Method}
\label{sec:method}
In this section we explain our data collection pipeline and class-agnostic lifting model.

\begin{figure*}
  \centering
  
  \includegraphics[width=\linewidth]{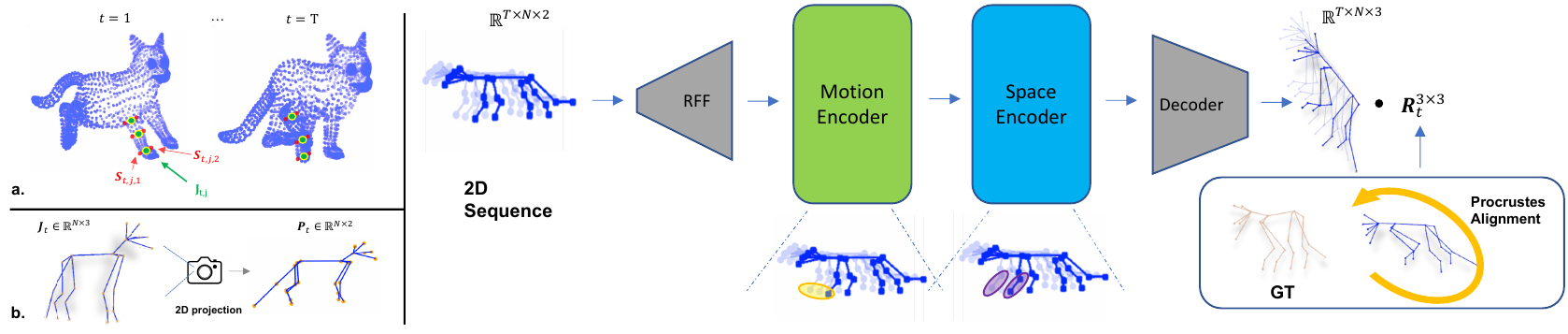}

   \caption{
   \textbf{Overview of our data pipeline and 3D lifting model.} The left side of the figure demonstrates (\textbf{a}) the process of calculating skeleton joints from animal mesh vertices, and (\textbf{b}) the projection of the those joints into 2D keypoints. The right side of the figure illustrates our lifting model at a high-level. The sequence of 2D input and temporal index is projected and passed through our motion encoder and space encoder layers. The spatio-temporal latent features are decoded into canonical 3D structures. The canonical structures are then aligned to the ground truth (GT) via procrustes-alignment for calculating the loss.
   }
   \label{fig:method}
\end{figure*}

\subsection{Dataset}
\label{sec:method_dataset}

We find a noticeable gap in available public datasets containing diverse 3D animal skeletons with realistic motion sequences. We aim to fill this gap by creating a new synthetic dataset, \textbf{AnimalSyn3D}, that builds on the mesh vertices and animation sequences provided by the DeformableThings4D \cite{li4DCompleteNonRigidMotion2021} dataset. We provide 3D skeleton labels for 13 animal categories, totalling 678 animation sequences with temporal correspondence across 48,384 frames. We provide statistics and examples of our dataset in the \cref{tab:supp_dataset}. We detail our data collection pipeline in following sections.

\subsubsection{Animal joints}
Given the vertices of a skin-tight mesh of an animal and an associated sequence of movement, our aim is to find the 3D locations of the anatomically-accurate skeleton joints of the animal. We define the locations of $K$ vertices and $N$ joints in 3D space throughout a sequence of $T$ frames as $\mathbf{V} \in \mathbb{R}^{T \times K \times 3}$ and $\mathbf{J} \in \mathbb{R}^{T \times N \times 3}$, respectively. Note that the number of mesh vertices, joints, and frames may vary across animals and sequences. We take inspiration from motion capture, where markers are attached to an object and used to estimate joint positions via triangulation. We strategically select a subset of $M$ vertices $\mathbf{S}_{t,j} = \bigl\{ \mathbf{S}_{t,j,1}, ..., \mathbf{S}_{t,j,M} \bigl\} \subset \mathbf{V}_t$ to be virtual markers, such that the mean of the markers will provide the location of a joint $j$ in frame $t$:

\begin{equation}
    \mathbf{J}_{t,j} = \frac{1}{m} \sum_{i=1}^{m}\mathbf{S}_{t,j,i}.
  \label{eq:joints}
\end{equation}

The selection of vertices for a subset $\mathbf{S}_{t,j}$ is guided by a visual inspection of $\mathbf{J}_{t,j}$ and the trajectory similarity of the chosen vertices. The bones of the animal are defined as an adjacency matrix $\mathbf{A}^{N \times N}$ containing the connections between joints. We decide the number of joints, their approximate locations, and their connections by reviewing the anatomical structure of the target animal.

\subsubsection{Non-linear optimisation}
The noise that is inherently present in our human annotation process occasionally results in the length of animal bones to change over time. We adopt an additional inverse-kinematics optimization procedure to refine the position of joints so that the bone lengths are consistent across time. We formulate the problem as solving for the pose angles $\theta \in \mathbb{R}^{T \times N \times 3}$ of forward kinematics for each joint $j$ in frame $t$:

\begin{equation}
    \mathbf{\widehat{J}}_{t,j} = f(\theta_{t,j}) = f(\theta_{t,p}) \cdot \\
    \begin{bmatrix}
        \mathcal{R}(\theta_{t,j}) & \mathbf{L}_{j,p} \\
        \mathbf{0} & \mathbf{1}
    \end{bmatrix},
  \label{eq:kinematics}
\end{equation}

where $p$ is the parent of joint $j$ in the kinematic chain, $\mathcal{R}$ transforms $\theta_{t,j}$ into a valid $\mathbf{R} \in SO(3)$ rotation matrix using Rodrigues' rotation formula, and $\mathbf{L}_{j,p}$ is the bone length between joints $j$ and $p$ in the first frame. Forcing the $\mathbf{L}$ translation vector to be from a single frame forces bone lengths to be the same for every frame.

We use gradient descent with the Adam \cite{kingmaAdamMethodStochastic2017a} optimizer to optimise the objective function

\begin{equation}
    \underset{\mathbf{\theta}}{\text{minimize}} \sum_{t=1}^{T} \sum_{j=1}^{N} \Big\| \mathbf{\widehat{J}}_{t,j} - \mathbf{J}_{t,j} \Big\|_2 + \lambda L_S,
  \label{eq:kinematics_objective}
\end{equation}

where ${L}_S$ is an additional smoothness regulariser:

\begin{equation}
    L_S = \Big\| \mathbf{\widehat{J}}_{t,j} - \mathbf{\widehat{J}}_{t-1,j} \Big\|_2.
  \label{eq:kinematics_objective_smooth}
\end{equation}

The inverse-kinematics optimization provides a new set of joints $\mathbf{\widehat{J}}_{t,j}$ that ensures consistent bone lengths throughout any complex sequence of movement.

\subsubsection{Perspective projection}
We define an initial camera pose to satisfy two conditions. The mean location of the animal throughout the sequence is at the center of the camera view, and all joints are within view for the entirety of the sequence. We randomly rotate the camera around the y-axis and project the points to the 2D camera plane. For our purpose we do this only once for each animation sequence, however this process can be used to obtain many different views of the animal throughout the sequence.

\subsection{Lifting model}

Given an input sequence of 2D skeletons $\mathbf{X} \in \mathbb{R}^{T \times J \times 2}$, where $T$ is the number of frames in the video and $J$ is the number of joints, our goal is to reconstruct the 3D skeletons $\mathbf{\widehat{Y}} \in \mathbb{R}^{T \times J \times 3}$ of the object.

\subsubsection{Keypoint features}
The attention mechanism of transformers is inherently permutation equivariant, such that inputs can be randomly permuted and the corresponding outputs will remain the same. We leverage this property to handle objects with different joint configurations. We utilise the masking mechanism of \cite{dabhi3DLFMLiftingFoundation2024} to overcome the technical challenge of training with different numbers of joints. Inputs are zero-padded up to the maximum number of joints in a mini-batch and a mask $\mathbf{M} \in \{ 0,1 \}^J$ is used to ignore padded joints. Each element in $\mathbf{M}$ is defined as:


\begin{equation}
    \mathbf{M}_i = 
        \begin {cases}
             1 & \text{if joint $i$ is present} \\
             0 & \text{otherwise}
        \end{cases}
  \label{eq:mask}
\end{equation}

We encode the 2D skeletons into $D$-dimensional features $\mathbf{F} \in \mathbb{R}^{T \times J \times D}$ using Random Fourier Features (RFF) \cite{zhengRobustPointCloud2023}. We additionally encode each 2D joint $(x, y)$ with its temporal location $t$ in the sequence. We thus compute the feature of an input $\mathbf{p} = [x, y, t]^T$ as:

\begin{equation}
    \phi(\mathbf{p}) = \sqrt{\frac{2}{D}} \Bigl[ \sin(\mathbf{W} \cdot \mathbf{p} + \mathbf{b}) ; \cos(\mathbf{W} \cdot \mathbf{p} + \mathbf{b}) \Bigr],
  \label{eq:rff}
\end{equation}

where $\mathbf{W} \in \mathbb{R}^{\frac{D}{2} \times 3}$ is sampled from a normal distribution $\mathcal{N}(0, I)$ and $\mathbf{b} \in \mathbb{R}^{\frac{D}{2}}$ is sampled from a uniform distribution $\mathcal{U}(0, \frac{1}{2\pi})$. We choose analytical RFF for its success in low-data and out-of-distribution (OOD) scenarios \cite{dabhi3DLFMLiftingFoundation2024,zhengRobustPointCloud2023}. We find that encoding the temporal position is beneficial for the motion encoder in capturing temporal dependencies.

\subsubsection{Motion encoder}
The motion encoder leverages multi-head self-attention (MHSA) to embed temporal context into the keypoint features. Although we choose MHSA for its ability to use information from all frames, its lack of an inductive bias makes it unsuitable for tasks with limited data. We argue that it is not necessary to consider all frames in a sequence because most of the useful information can be found in nearby frames. We explicitly impose this inductive bias into the MHSA blocks by applying a binary mask to the intermediate attention maps. We define the mask $\mathbf{Z} \in \mathbb{R}^{T \times T}$ for a joint at time $t$ as:

\begin{equation}
    \mathbf{Z}_{t,i} = 
        \begin {cases}
             1 & \text{for } t-\alpha \leq i \leq t+\alpha \\
             0 & \text{otherwise}
        \end{cases}
  \label{eq:constraint}
\end{equation}

where $\alpha$ is a hyper-parameter controlling the number of frames before and after $t$ that can contribute information during attention.

Given a set of spatial features $\mathbf{F} \in \mathbb{R}^{T \times J \times D}$ as input to our motion encoder, we matrix transpose to get $\mathbf{F}_M \in \mathbb{R}^{J \times T \times D}$. We first apply linear projections to obtain queries $\mathbf{Q}$, keys $\mathbf{K}$, and values $\mathbf{V}$ for each head $h$:

\begin{equation}
    \mathbf{Q}^{(h)} = \mathbf{F} \mathbf{W}_{Q}^{(h)}, \mathbf{K}^{(h)} = \mathbf{F} \mathbf{W}_{K}^{(h)}, \mathbf{V}^{(h)} = \mathbf{F} \mathbf{W}_{V}^{(h)},
  \label{eq:projection}
\end{equation}

where $\mathbf{Q}^{(h)}, \mathbf{K}^{(h)}, \mathbf{V}^{(h)} \in \mathbb{R}^{N, T, \frac{D}{H}}$ for a total of $H$ heads. We apply our temporal mask $\mathbf{Z}$ before the non-linear softmax operation in each head:

\begin{equation}
    \text{head}_{h} = \text{softmax}(\frac{\mathbf{Q}^{(h)} (\mathbf{K}^{(h)})^{\mathbf{T}}}{\sigma} \times \mathbf{Z}) \cdot \mathbf{V}^{(h)},
  \label{eq:attn_head}
\end{equation}

where $\sigma$ is a scaling factor. Finally, each head is concatenated and projected:

\begin{equation}
    \mathbf{F}_M = \text{MHSA}_{\alpha}(\mathbf{F}_M) = [head_1;...;head_h]\mathbf{W_P}.
  \label{eq:attn}
\end{equation}

We follow the standard procedure of applying a residual connection and layer normalisation before obtaining the final output of one windowed-MHSA layer. We stack $P$ of these layers with residual connections to create our motion encoder.

\subsubsection{Space encoder}
The space encoder uses the features from our motion encoder to model the relationships among joints in a single frame. We found the hybrid graph-based approach of \cite{dabhi3DLFMLiftingFoundation2024} to perform favorably for our task. Given the transposed motion features $\mathbf{F}_{M} \in \mathbb{R}^{T \times J \times D}$, a single space-layer has two simultaneous processing streams, one for capturing the local connectivity between joints $G_\text{local}$, and another for capturing global connectivity $G_\text{global}$. These two streams are concatenated and projected to provide an output of spatial features containing a combination of both streams:
\begin{equation}
    \mathbf{F}_S = \text{MLP}([G_\text{local}(\mathbf{F}_M, \mathbf{A}); G_{\text{global}}(\mathbf{F}_M)]). \\
  \label{eq:space}
\end{equation}
The local graph-attention $G_\text{local}$ utilises an adjacency matrix of the joint connections $\mathbf{A} \in \mathbb{R}^{J\times J}$ to model the object connectivity. A layer normalisation and skip connection is applied to produce the final output of a layer. 
As with our motion encoder, we stack $O$ layers with residual connections between them.

\subsubsection{Decoder and Procrustes-based loss}
Lastly, given the latent spatial features $\mathbf{F}_S$, we use an MLP to decode the predicted 3D structures of the object in a canonical 3D space
\begin{equation}
    \mathbf{\widehat{Y}}^\text{canon} = \text{MLP}(\mathbf{F}_S).
  \label{eq:decoder}
\end{equation}

We align each canonical prediction $\mathbf{\widehat{Y}}^\text{canon} \in \mathbb{R}^{T \times J \times 3}$ with the ground truth  $\mathbf{Y}$ via a Procrustes alignment method which solves for the optimal rotation $\mathbf{\tilde{R}}_t$ individually for each frame $t$ as:
\begin{equation}
    \underset{\mathbf{R}_{t} \in SO(3)}{\text{minimize}}~ \Big\| \mathbf{Y}_t - \mathbf{\widehat{Y}}^{\text{canon}}_{t} \mathbf{R}_t \Big\|_2.
  \label{eq:procrustes_rotation}
\end{equation}
In practice, we use Singular Value Decomposition (SVD) to solve this optimisation problem. To ensure that $\mathbf{\tilde{R}}_t$ belongs to the special orthogonal group $SO(3)$, we enforce $\det(\mathbf{\tilde{R}}_t)$$=+1 $. This step is crucial for mitigating reflection ambiguity in our predictions. 

The resulting $\mathbf{\tilde{R}}_t$ is used to align our canonical predictions with the ground truth. We additionally scale the predictions relative to the ground truth using a scaling factor $\mathbf{s} \in \mathbb{R}^{T}$:

\begin{equation}
    \mathbf{\widehat{Y}}_t = \mathbf{s}_t \cdot (\mathbf{\widehat{Y}}^{\text{canon}}_{t} \mathbf{\tilde{R}}_t)
  \label{eq:procrustes}
\end{equation}

\subsubsection{Loss function}
With our predictions now aligned with the ground truth, we can compute our loss. We compute the Mean Squared Error (MSE) of the 3D points along with an additional velocity error $\mathcal{L}_\text{vel}$:

\begin{equation}
    \mathcal{L}_{\text{total}} = \sum_{t=1}^{T} \sum_{j=1}^{J} \Big\| \mathbf{Y}_{t,j} - \mathbf{\widehat{Y}}_{t,j} \Big\|_2 + \lambda \mathcal{L}_\text{vel},
  \label{eq:loss}
\end{equation}

\begin{equation}
    \mathcal{L}_\text{vel} = \sum_{t=2}^{T} \sum_{j=1}^{J} \Big\| (\mathbf{Y}_{t,j} - \mathbf{Y}_{t-1,j}) - (\mathbf{\widehat{Y}}_{t,j} - \mathbf{\widehat{Y}}_{t-1,j}) \Big\|_2.
  \label{eq:vel_loss}
\end{equation}

We use the scalar $\lambda$ to weight the velocity loss.

\section{Experiments}
\label{sec:experiments}

\definecolor{Gray}{gray}{0.85}
\def\thickhline{\noalign{\hrule height2.0pt}}
\begin{table*}[t]
  \setlength\tabcolsep{4.0pt}
  \centering
  {\small{
  \begin{tabular}{@{}l|ccccccccccccc|>{\columncolor{Gray}}c@{}}
    \toprule
    Method & Bear & Buck & Bunny & Chicken & Deer & Dog & Elk & Fox & Moose & Puma & Rabbit & Raccoon & Tiger & Avg \\
    \hline
    MotionBERT & 94.5 & 208.1 & \textbf{16.7} & 108.2 & 200.7 & 50.1 & 267.4 & 40.6 & 189.2 & 254.4 & 30.7 & 77.4 & 211.8 & 134.6 \\
    3D-LFM & 47.6 & 158.2 & 23.2 & 92.3 & 156.8 & 53.9 & 147.8 & 22.2 & 274.7 & 163.4 & 37.8 & 70.0 & 165.4 & 108.7 \\
    Ours & \textbf{29.2} & \textbf{128.4} & 17.1 & \textbf{60.8} & \textbf{57.3} & \textbf{32.8} & \textbf{103.1} & \textbf{14.2} & \textbf{97.9} & \textbf{93.2} & \textbf{19.0} & \textbf{44.5} & \textbf{90.8} & \textbf{60.6} \\
    \hline
    \hline
    MotionBERT & 90.7 & 198.1 & 16.0 & 99.0 & 195.5 & 45.8 & 246.8 & 39.9 & 170.9 & 235.0 & 28.6 & 74.9 & 203.2 & 126.5 \\
    3D-LFM & 27.9 & 108.3 & 12.2 & 86.3 & 75.0 & 33.3 & 103.0 & 16.2 & 119.7 & 119.3 & 21.2 & 57.6 & 107.7 & 68.3 \\
    Ours & \textbf{26.7} & \textbf{107.3} & \textbf{11.2} & \textbf{54.2} & \textbf{50.9} & \textbf{27.9} & \textbf{86.1} & \textbf{12.4} & \textbf{81.6} & \textbf{85.9} & \textbf{15.4} & \textbf{42.8} & \textbf{79.8} & \textbf{52.5} \\
    \hline
    \hline
    MotionBERT & 3.2 & \textbf{11.0} & \textbf{1.1} & 3.9 & 6.9 & \textbf{2.6} & 10.4 & 1.3 & 17.8 & 9.8 & 2.1 & 4.0 & 9.8 & 6.5 \\
    3D-LFM & 7.6 & 29.0 & 3.4 & 8.4 & 26.3 & 8.4 & 26.5 & 3.8 & 43.4 & 27.3 & 6.7 & 12.3 & 30.4 & 18.0 \\
    Ours & \textbf{2.5} & 12.1 & 1.3 & \textbf{3.4} & \textbf{5.9} & 2.9 & \textbf{9.3} & \textbf{1.2} & \textbf{12.5} & \textbf{8.9} & \textbf{2.0} & \textbf{3.6} & \textbf{9.1} & \textbf{5.7} \\
    \bottomrule
  \end{tabular}
  }}
  \caption{\textbf{Quantitative comparison of 2D to 3D lifting with 13 animals.} We report, in millimeters, the Sequence-Aligned MPJPE (top), Frame-Aligned MPJPE (middle), and Sequence-Aligned MPVE (bottom), see \cref{subsec:eval_protocol} for details of these evaluation metrics. Our approach (Ours) outperforms existing state-of-the-arts with significant gap across multiple animal categories.
  }
  \label{tab:comparison_animal}
\end{table*}

\begin{table}
  \setlength\tabcolsep{3.0pt}
  \centering
  {\small{
  \begin{tabular}{@{}c|c|ccc@{}}
    \toprule
    Method & MC & FA-MPJPE$\downarrow$ & SA-MPJPE$\downarrow$ & SA-MPVE$\downarrow$ \\
    \midrule
    \multirow{2}{*}{MotionBERT} & - & 176.0 & 199.9 & 9.78 \\
                                & \checkmark & \textbf{126.5} & \textbf{134.6} & \textbf{6.5} \\
    \midrule
    \multirow{2}{*}{3D-LFM} & - & 89.2 & 126.4 & 20.5 \\
                            & \checkmark & \textbf{68.3} & \textbf{108.7} & \textbf{18.0} \\
    \midrule
    \multirow{2}{*}{Ours}   & - & 105.4 & 128.4 & 11.0 \\
                            & \checkmark & \textbf{52.5} & \textbf{60.6} & \textbf{5.7} \\
    \bottomrule
  \end{tabular}
  }}
  \caption{ 
  \textbf{Quantitative comparison between multi-category (MC) and single-category training.} We use a \checkmark for models trained with multi-category training. Each method benefits from multi-category training. See \smat{} for a breakdown of per-animal results.
  }
  \label{tab:single_vs_multi}
\end{table}

We evaluate our method on various animal categories to assess its performance and generalisation properties. Comparative analyses are with recent state-of-the-art video (MotionBERT~\cite{zhuMotionBERTUnifiedPerspective2023}) and single-frame (3D-LFM~\cite{dabhi3DLFMLiftingFoundation2024}) lifting models on various animal categories and motion sequences. 

\paragraph{Datasets}

We use the AnimalSyn3D dataset as described in \cref{sec:method_dataset}. The 2D keypoints provided by off-the-shelf pose detectors~\cite{xuViTPoseSimpleVision2022} are inherently noisy due to factors such as lighting conditions and image quality. To simulate these conditions, we synthetically perturb the 2D keypoints with an additive Gaussian noise, which corresponds to a $3$-pixel error on average. We present results for non-noisy data (\cref{tab:supp_clean}) and also provide comparisons on 3D human pose estimation in the \smat{} (\cref{tab:supp_human}), although human-specific lifting is not the focus of our work. We normalise 2D keypoints and 3D labels to $[-1, 1]$, after scaling the 3D labels following existing works on 3D human pose estimation~\cite{zhuMotionBERTUnifiedPerspective2023}. We split the data for training by randomly selecting 80\% of the animation sequences for each animal, with the remaining animations withheld for testing.

\paragraph{Evaluation protocols}\label{subsec:eval_protocol}
Previous video lifting methods \cite{zhuMotionBERTUnifiedPerspective2023,chenAnatomyaware3DHuman2021,hossainExploitingTemporalInformation2018a,liuAttentionMechanismExploits2020} evaluate the non-rigid structure and motion of their approach by calculating the mean per-joint position error (MPJPE) directly with the ground truth in the \textit{camera space}. However, our model and 3D-LFM make predictions in a \textit{canonical space} that requires the alignment of each frame to the ground truth. As such, we instead use the standard \textit{per-frame} Procrustes-aligned MPJPE metric and refer to it as the frame-aligned MPJPE \textbf{(FA-MPJPE)} for brevity. Additionally, we compose a metric to measure the relative motion error in a video sequence.

\paragraph{Sequence-Aligned MPJPE}
We formulate the sequence-aligned MPJPE \textbf{(SA-MPJPE)} as solving for a \textit{single} rotation matrix $\mathbf{R} \in SO(3)$ to align the 3D predictions $\mathbf{\widehat{Y}}$ and ground truth $\mathbf{Y}$ for all $T$ frames in a sequence. This is in contrast to the SA-MPJPE that aligns each individual frame in a sequence by solving for $T$ rotation matrices. After alignment, we compute the MSE to produce the final error value. Let us specify $\mathbf{\widehat{Y}}, \mathbf{Y} \in \mathbb{R}^{T \times J \times 3}$, for $J$ joints, to define our metric as

\begin{equation}
    \underset{\mathbf{R}}{\text{minimize}} \sum_{t=1}^{T} \Big\| \mathbf{Y}_t - \mathbf{\widehat{Y}}_t\mathbf{R} \Big\|_2.
  \label{eq:sampjpe}
\end{equation}

Compared to \cref{eq:procrustes_rotation}, we are instead solving for a single, global $\mathbf{R}$. Our metric thus captures any error in the motion that occurs between subsequent frames of a sequence.

Lastly, we are able to report the commonly used mean per-joint velocity error (MPVE) after performing the global sequence alignment. For clarity in our comparisons, we refer to this as \textbf{SA-MPVE}.

\paragraph{Implementation details}
Here we provide important implementation details of our method and refer the reader to the \smat{} for further details. We construct batches of 32 sequences with 48 frames per sequence. We train on a total of 871 video sequences and evaluate on a separate set of 199 unseen sequences across all 13 animal categories. Inputs are zero-padded up to the maximum of 29 joints that occur in dataset. A layer size of $P=4$ is chosen for the motion encoder and $O=12$ for the space encoder. The hidden-dimension size $D$ is 256. Experiments were conducted on a single NVIDIA A100 GPU.

MotionBERT uses human-specific semantic knowledge, making it unsuited for object-agnostic lifting. To provide a fair comparison, we do not modify the proposed architecture and instead apply two alternative alterations. We first set the number of learned positional embeddings to the maximum amount of joints seen in the dataset, allowing it to handle all animal rigs. We also randomly permute the 2D inputs during training and testing to simulate a real object-agnostic scenario. This is required so that the positional embeddings are not learning dataset-specific skeleton semantics. We ensure that joints being permuted retain their temporal correspondence over a sequence. 

\subsection{Object-agnostic lifting}\label{subsec:multi-cat}
\cref{tab:comparison_animal} demonstrates the effectiveness of our method compared to MotionBERT and 3D-LFM. We outperform both methods across all three metrics and nearly every animal. Notably, our approach achieves 45\% lower SA-MPJPE and 70\% lower SA-MPVE compared to 3D-LFM, demonstrating that our predicted 3D motion has more accurate and smoother movement, while also preserving high-fidelity object structure (FA-MPJPE). This substantial performance gap highlights the critical role of the motion encoder in capturing the temporal relationship of joints. While MotionBERT is a spatio-temporal approach, it has no inductive bias to assist with handling scarce data and multi-category training. \cref{fig:deer} qualitatively demonstrates the predictions of our method compared to 3D-LFM. 

\begin{figure}[t]
  \centering
  
  \subfloat[3D-LFM]{
    \includegraphics[width=0.45\linewidth]{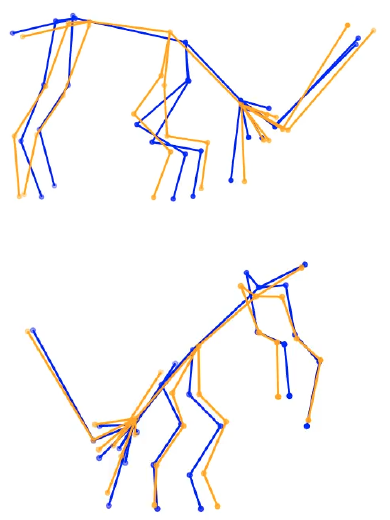}
    \label{fig:deer_3dlfm}
  }
  \hfill
  \vrule
  \hfill
  \subfloat[Ours]{
    \includegraphics[width=0.45\linewidth]{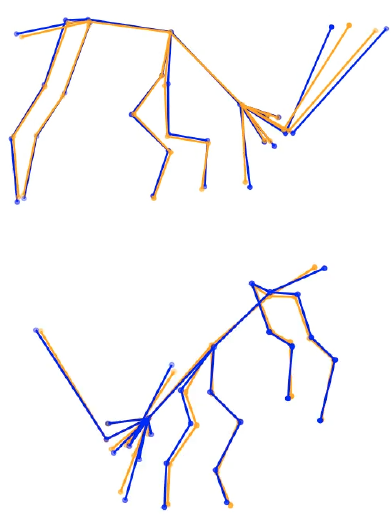}
    \label{fig:deer_ours}
  }
   \caption{\textbf{Quantitative comparison on a \textit{Deer} sequence from two different views}: Our method provides significantly more accurate 3D predictions. In this visualization, blue represents the predicted 3D points whereas the orange denotes the ground truth.}
   \label{fig:deer}
\end{figure}

\paragraph{Single- vs multi-category}
We evaluate performance on single-category training and compare it to their performance on multi-category training. For single-category, we train each approach from scratch using data specific to a single animal category. This process is repeated for all $13$ animals, and the mean error is reported in \cref{tab:single_vs_multi}. All methods show significant improvement when performing multi-category training as opposed to single-category, achieving at least a 25\% reduction in FA-MPJPE, with our approach achieving a 50\% reduction. This improvement highlights the advantage of unified learning across a vast spectrum of object categories, particularly in scenarios where the training data is small and unbalanced.

\paragraph{Robustness to occlusion}
To evaluate the robustness of 3D-LFM and our approach to occlusion scenarios, we trained both models by randomly masking $10$\% of all 2D keypoints within a frame and tasking the models with recovering the 3D locations of the missing joints. As shown in \cref{fig:tiger}, our method demonstrates superior robustness while 3D-LFM struggles. Even in an extreme case of 60\% occlusion we continue to see legible predictions from our method; see \cref{fig:robustness_chart} in the \smat{} for a qualitative result.

\subsection{OOD generalization}

\paragraph{Unseen objects} We perform a $13$-fold analysis, where each fold involves holding out one animal category from the original dataset during training. For example, the \textit{bunny} category is excluded from the training data and used to evaluate the generalization capability of each method. As shown in \cref{fig:ood_graph} (left), our approach demonstrates superior OOD generalization when handling unseen animal categories by outperforming existing methods by a significant margin. We present a qualitative reconstruction for a \textit{bunny} instance in \cref{fig:rabbit} and refer the reader to \cref{tab:ood_animal} in the \smat{} for tabulated results.

\vspace{-1em}

\paragraph{Rig transfer} 
When lifting an unseen animal in the wild, we may encounter an animal with a more complex structure than seen during training. We showcase our ability to generalize to an unseen animal with an unseen number of joints. While MotionBERT is limited to rigs with the same or fewer joints as those seen during training, our method can handle \textit{any} number of joints. We train on animal rigs with $27$ or fewer joints and test on two \textit{unseen} animals with $29$ joints (deer and moose). As shown in \cref{fig:ood_graph} (right), while both 3D-LFM and our approach are impacted by the difficulty of the task, our approach outperforms 3D-LFM by at least 40\% in all metrics. We show the tabulated results in the \smat{} (\cref{tab:ood_animal_joints}).

\begin{figure}
    \centering
    \includegraphics[width=0.49\textwidth]{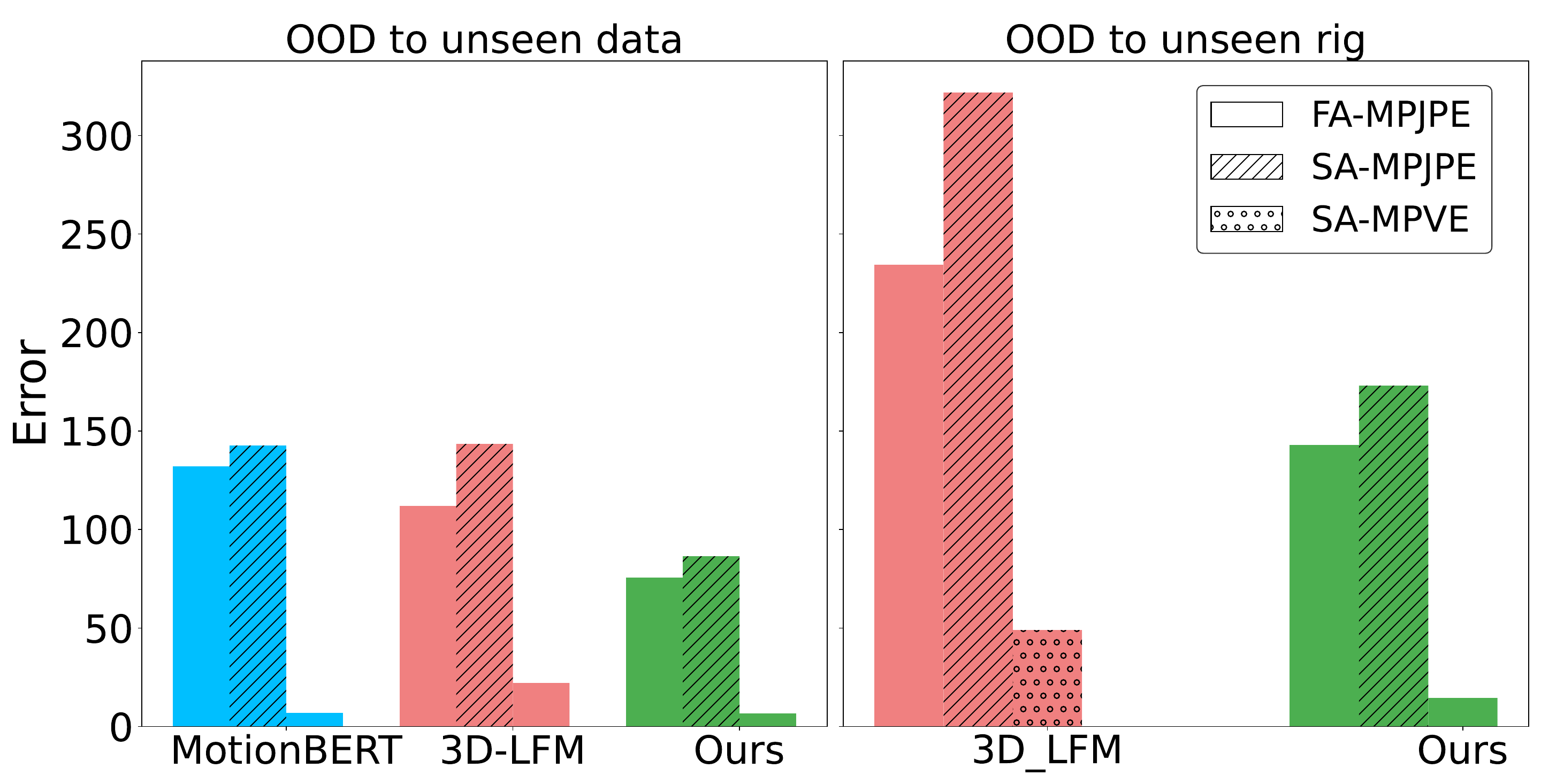}
    \caption{\textbf{OOD generalization}. \emph{OOD to unseen data} (left): We perform a 13-fold evaluation to assess each method's ability to handle unseen animal categories. \emph{OOD to an unseen category and rig} (right): Note that MotionBert is constrained to rigs with the same or fewer joints as those seen during training and hence cannot handle unseen rigs with more joints. Our method can handle generalization to both unseen category and unseen rig more effectively. }
    \label{fig:ood_graph}
\end{figure}

\begin{figure}[t]
  \centering
  
  \subfloat[3D-LFM]{
    \includegraphics[width=0.45\linewidth]{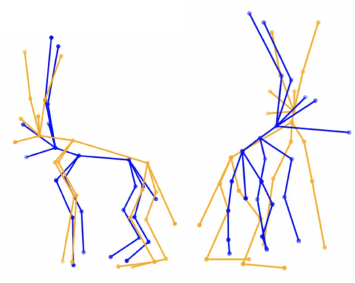}
    \label{fig:3dlfm_rabbit}
  }
  \hfill
  \vrule
  \hfill
  \subfloat[Ours]{
    \includegraphics[width=0.45\linewidth]{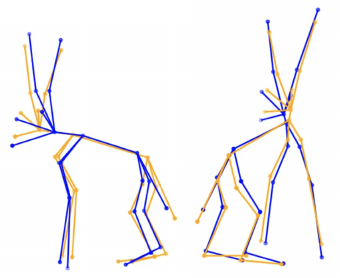}
    \label{fig:ours_rabbit}
  }
   \caption{\textbf{OOD generalization on an unseen \textit{Bunny} category from two different views}: Our method provides significantly more accurate 3D predictions compared to 3D-LFM. In this visualization, blue represents the predicted 3D points whereas the orange denotes the ground truth. }
   \label{fig:rabbit}
\end{figure}
\begin{figure}[t]
  \centering
  
  \subfloat[3D-LFM]{
    \includegraphics[width=0.45\linewidth]{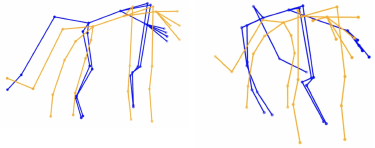}
    \label{fig:3dlfm_tiger}
  }
  \hfill
  \vrule
  \hfill
  \subfloat[Ours]{
    \includegraphics[width=0.45\linewidth]{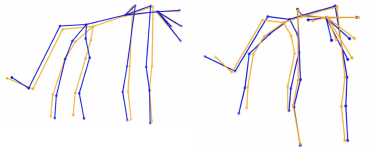}
    \label{fig:ours_tiger}
  }

   \caption{\centering
   \textbf{A comparison of robustness when 10\% of a tiger is occluded.} We include the average frame-aligned MPJPE across all animals for each method. Two views of the object are shown.}
   \label{fig:tiger}
\end{figure}

\subsection{Ablations}
In this section we ablate the important building blocks of our approach. We first highlight the importance of our spatio-temporal approach as opposed to a spatial-only approach. Then, we demonstrate the improvement gained from our temporal-proximity inductive bias. We go on to compare strategies for encoding the position of a joint in time. Lastly, we ablate the importance of our procrustes-based training.
\paragraph{Time vs. space}
We evaluate the significance of information sharing across time facilitated by our proposed motion encoder. To isolate the impact of time, we create a space-only variant of our model by replacing the motion encoder with additional space blocks such that the number of parameters remains similar. As shown in \cref{tab:ablate_time}, modelling temporal dependencies is crucial for enhancing 2D-3D lifting performance, particularly in terms of per-sequence reconstruction accuracy (SA-MPJPE) and smoothness (SA-MPVE).

\begin{table}
  \centering
  {\small{
  \begin{tabular}{@{}c|ccc@{}}
    \toprule
    Model & FA-MPJPE$\downarrow$ & SA-MPJPE$\downarrow$ & SA-MPVE$\downarrow$ \\
    \midrule
    Space & 67.8 & 99.9 & 13.1 \\
    Time & \textbf{57.9} & \textbf{66.5} & \textbf{5.6} \\
    \bottomrule
  \end{tabular}
  }}
  \caption{
  \textbf{Ablation of time vs. space.}}
  \label{tab:ablate_time}
\end{table}

\paragraph{Constrained temporal attention}
We observe that applying our inductive bias to restrict information sharing enhances the optimization of our model. We conduct our ablation by progressively increasing the window size $\alpha$. We apply $\alpha$ according to \cref{eq:constraint}.
\cref{tab:ablate_constrained} shows the impact of different $\alpha$ on performance, averaged over five statistical runs. Overall, we observe a trade-off between the accuracy of the 3D structure and the accuracy of 3D motion. In our case, we identify $\alpha = 8$ as optimal, which we used for all of our other experiments.

\begin{table}
  \centering
  {\small{
  \begin{tabular}{@{}c|ccc@{}}
    \toprule
    $\alpha$ & FA-MPJPE $\downarrow$ & SA-MPJPE $\downarrow$ & SA-MPVE $\downarrow$ \\
    \midrule
    2 & $53.9_{\pm1.12}$ & $65.5_{\pm0.96}$ & $7.0_{\pm0.02}$ \\
    4 & $54.4_{\pm1.13}$ & $64.9_{\pm2.78}$ & $6.4_{\pm0.01}$ \\
    8 & $\mathbf{52.7}_{\pm0.19}$ & $\mathbf{61.4}_{\pm0.40}$ & $5.8_{\pm0.00}$ \\
    16 & $57.1_{\pm0.93}$ & $65.6_{\pm1.70}$ & $5.7_{\pm0.02}$ \\
    - & $57.9_{\pm0.33}$ & $66.5_{\pm0.88}$ & $\mathbf{5.6}_{\pm0.01}$ \\
    \bottomrule
  \end{tabular}
  }}
  \caption{\textbf{Ablation of our inductive bias}. We use $-$ to denote no constraint being applied. We report the averages over $5$ independent runs.}
  \label{tab:ablate_constrained}
\end{table}



\paragraph{Analytical temporal embedding}
Here we ablate our use of analytical RFF to encode temporal information. In \cref{tab:ablate_emb}, we show that it is beneficial to use an analytical embedding over a learned embedding. Analytical RFF provides a significant performance increase over a learned embedding, in our case it seems to result from a scarcity of data.

\begin{table}
  \centering
  {\small{
  \begin{tabular}{@{}c|ccc@{}}
    \toprule
    Temporal Embedding & FA-MPJPE$\downarrow$ & SA-MPJPE$\downarrow$ & SA-MPVE$\downarrow$ \\
    \midrule
    - & $55.5_{\pm2.29}$ & $64.8_{\pm3.04}$ & $5.9_{\pm0.03}$ \\
    Learned & $54.0_{\pm1.13} $& $64.1_{\pm1.24}$ & $6.1_{\pm0.02}$ \\
    Analytical & $\mathbf{52.7}_{\pm0.25}$ & $\mathbf{61.4}_{\pm0.27}$ & $\mathbf{5.8}_{\pm0.01}$ \\
    \bottomrule
  \end{tabular}
  }}
  \caption{
   \textbf{Ablation of temporal embedding strategies.} We report the averages over $5$ independent runs.
  }
  \label{tab:ablate_emb}
\end{table}

\paragraph{Procrustes-based training}
Similar to 3D-LFM, we find it useful to train our model with a Procrustes-based loss, as shown in \cref{tab:ablate_Procrustes}.
Allowing the model to focus solely on learning object structure significantly enhances the predicted 3D structure and benefits overall 3D motion accuracy. Interestingly, we observe that we are able to perform well without the Procrustes-based loss, even outperforming 3D-LFM across all metrics. This offers an interesting insight into future work involving our method for practical applications that require implicitly predicted camera rotation.

\begin{table}
  \centering
  {\small{
  \begin{tabular}{@{}c|c|ccc@{}}
    \toprule
    Model & Procrustes & FA-MPJPE$\downarrow$ & SA-MPJPE$\downarrow$ & SA-MPVE$\downarrow$ \\
    \midrule
    3D-LFM & - & 83.2 & 97.3 & 25.0 \\
    3D-LFM & \checkmark & 68.3 & 108.7 & 18.0 \\
    Ours & - & 62.2 & 74.9 & 7.7 \\
    Ours & \checkmark & \textbf{52.7} & \textbf{61.4} & \textbf{5.6} \\
    \bottomrule
  \end{tabular}
  }}
  \caption{\textbf{Procrustes vs. non-Procrustes training}. Our spatio-temporal 2D-3D lifting approach with Procrustean alignment surpasses 3D-LFM.}
  \label{tab:ablate_Procrustes}
\end{table}

\section{Conclusion}
\label{sec:conclusion}

In this work, we introduced an object-agnostic 3D lifting model that leverages a temporal inductive bias for temporal sequences. Our approach sets a new benchmark in lifting performance for object categories with limited available data. The model's ability to generalize across unseen object categories and rigs shows its versatility and robustness, even in challenging scenarios involving noise and occlusions. In addition to these contributions, we have introduced a new synthetic dataset, AnimalSyn3D, designed to stimulate further research in class-agnostic 3D lifting models. This work aims to pave the way for more generalized and efficient 3D reconstruction methods that can be applied across diverse real-world applications.


\section{Acknowledgements}
\label{sec:acknowledgements}

We would like to thank Angel Allen for contributing the dog and raccoon to the dataset, Hemanth Saratchandran for polishing the mathematical notations, Lachlan Mares and Stefan Podgorski for notes during drafting, and Chee-Keng Ch'ng for helpful discussions throughout the project.
{
    \small
    \bibliographystyle{ieeenat_fullname}
    \bibliography{main}
}
}{} 
\ifthenelse{\boolean{show_supp}}{
\clearpage
\setcounter{page}{1}
\maketitlesupplementary

\setcounter{section}{0}
\renewcommand\thesection{\Alph{section}}

\section{Extra training details}
\label{sec:supp_implementation}
We train our model with the Adam \cite{kingmaAdamMethodStochastic2017a} optimizer using $\text{lr}=10^{-4}$, weight decay of $10^{-6}$, $\beta_1=0.9$, and $\beta_2=0.999$. We set the velocity loss term $\lambda=5000$. We train for a total of $1000$ epochs and use the model with the lowest validation error out of all epochs.

\section{Dataset details}
\label{sec:supp_dataset}

\subsection{Statistics}
An overview of our dataset can be seen in \cref{tab:supp_dataset}. We showcase the number of joints, animations, and frames for each animal category. Our dataset contains a variety of different categories and rigs to facilitate class-agnostic training and evaluation.

\subsection{Examples}
We provide an additional \textbf{dataset\_examples.mp4} video in the supplementary zip file for viewing the 3D skeletons for some animals.

\section{Additional results}

\subsection{Examples}
We provide an additional \textbf{prediction\_examples.mp4} video in the supplementary zip file for viewing some predictions over a series of video sequences. We provide a side-by-side comparison between our approach and 3D-LFM. We also provide examples of out-of-distribution predictions. We believe that these videos provide a more comprehensive comparison than that which can be obtained with 2D images.

\subsection{Ablating 2D noise}
\label{sec:supp_clean}
Alongside \cref{tab:comparison_animal} in the main paper, we provide additional experiments on our dataset where we do not apply synthetic noise to the 2D inputs. \cref{tab:supp_clean} shows that the improvements provided by our approach over existing methods is not restricted to situations with noisy 2D poses.

\subsection{Human3.6M benchmark}
\label{sec:supp_human}
We provide comparisons on the Human3.6M \cite{ionescuHuman36MLarge2014} benchmark containing 3.6 million video frames of real humans performing simple tasks in a controlled indoor environment. Following previous lifting works \cite{zhuMotionBERTUnifiedPerspective2023,brau3DHumanPose2016}, we train on subjects 1, 5, 6, 7, and 8, and hold out subjects 9 and 11 for testing. We use the noisy 2D skeletons provided by \cite{zhuMotionBERTUnifiedPerspective2023} that were obtained using Stacked Hourglass Networks \cite{newellStackedHourglassNetworks2016}. Every fifth frame is used for all experiments with no significant degradation in performance. We additionally compare with the original MotionBERT \cite{zhuMotionBERTUnifiedPerspective2023}, denoted with $\dagger$ in \cref{tab:supp_human}, that uses human-specific augmentation during training and human-specific semantic correspondences. We find that, when there is an abundance of data for a single object, leveraging object-specific information is preferred to class-agnostic training. \cref{tab:supp_human} further demonstrates the benefit of our approach over the current SOTA class-agnostic method \cite{dabhi3DLFMLiftingFoundation2024}. We outperform 3D-LFM on each metric, translating to improved 3D object structure and motion consistency across entire sequences. The class-agnostic MotionBERT outperforms both our approach and 3D-LFM, likely due to the having twice as many parameters and an architecture that was specifically designed for large-scale human data.

\subsection{Per-animal multi-category}
\label{sec:supp_category}
Here we provide the per-animal results that correspond to \cref{tab:single_vs_multi} in the main paper. \cref{tab:supp_multicategory} demonstrates the usefulness of training with all categories at once instead of specializing to a single category. We find that this holds true for our model across all animal categories. The same is seen for MotionBERT and 3D-LFM with exception of the chicken category, where it is sometimes slightly better to do chicken-only training.

\subsection{Out-of-distribution categories and rigs}
\label{sec:supp_ood}
Here we provide tabulated results for our out-of-distribution (OOD) experiments found in Sec. 4.2 in the main paper. We show in \cref{tab:ood_animal,tab:ood_animal_joints} that our model does indeed maintain superior OOD performance across all metrics compared to existing methods. In the case of unseen number of joints (\cref{tab:ood_animal_joints}), we cannot evaluate MotionBERT as it is unable to handle a number of joints that is more than the maximum seen during training.

\subsection{Extreme occlusion}
\label{sec:supp_occlusion}
We provide a visual comparison between our approach and 3D-LFM when there is an extreme (60\%) occlusion of the object. \cref{fig:bear} demonstrates the robustness of our approach in this scenario. While 3D-LFM fails to properly reconstruct the 3D object structure (\cref{fig:3dlfm_bear}) for even a single frame, our method is capable of maintaining high-fidelity reconstruction (\cref{fig:ours_bear}). We show FA-MPJPE results for intermediate occlusion levels in \cref{fig:robustness_chart}.


\begin{table*}[t]
  \setlength\tabcolsep{4.0pt}
  \centering
  {\small{
  \begin{tabular}{@{}l|ccccccccccccc|c@{}}
    \toprule
    & Bear & Buck & Bunny & Chicken & Deer & Dog & Elk & Fox & Moose & Puma & Rabbit & Raccoon & Tiger & Total \\
    \midrule
    Animations & 67 & 42 & 45 & 7 & 56 & 65 & 67 & 37 & 59 & 68 & 45 & 54 & 66 & 678 \\
    Frames & 4,464 & 3,168 & 3,072 & 432 & 3,648 & 4,128 & 5,328 & 2,304 & 3,792 & 5,808 & 3,072 & 4,176 & 4,992 & 48,384 \\
    Joints & 21 & 27 & 25 & 19 & 29 & 22 & 26 & 26 & 29 & 26 & 25 & 28 & 27 & 330 \\
    \bottomrule
  \end{tabular}
  }}
  \caption{
  \textbf{An overview of our synthetic dataset}. We measure the total number of animation sequences and frames for each animal. We also provide the number of joints that constitute each animal.
  }
  \label{tab:supp_dataset}
\end{table*}

\begin{table*}[t]
  \setlength\tabcolsep{4.0pt}
  \centering
  {\small{
  \begin{tabular}{@{}l|cccccccccccccc@{}}
    \toprule
    Method & Bear & Buck & Bunny & Chicken & Deer & Dog & Elk & Fox & Moose & Puma & Rabbit & Raccoon & Tiger & Avg \\
    \midrule
    MotionBERT & 94.9 & 216.1 & 17.7 & 110.7 & 198.4 & 49.1 & 233.1 & 38.0 & 161.8 & 257.4 & 28.1 & 83.1 & 220.8 & 131.5 \\
    3D-LFM & 49.8 & 151.2 & 23.8 & 106.7 & 151.3 & 53.6 & 149.8 & 21.5 & 277.0 & 157.4 & 40.4 & 62.3 & 162.9 & 108.0 \\
    Ours & \textbf{28.7} & \textbf{125.4} & \textbf{15.8} & \textbf{80.4} & \textbf{59.4} & \textbf{31.6} & \textbf{122.4} & \textbf{13.7} & \textbf{98.2} & \textbf{93.7} & \textbf{17.6} & \textbf{45.4} & \textbf{93.1} & \textbf{63.5} \\
    \midrule
    \midrule
    MotionBERT & 90.9 & 205.3 & 17.0 & 98.6 & 189.9 & 45.7 & 211.7 & 36.4 & 143.3 & 237.5 & 26.3 & 80.7 & 210.9 & 122.6 \\
    3D-LFM & 31.3 & \textbf{100.4} & 12.7 & 97.6 & 67.4 & 33.8 & 104.4 & 15.3 & 113.5 & 114.3 & 26.4 & 49.6 & 107.1 & 67.1 \\
    Ours & \textbf{26.3} & 108.9 & \textbf{10.6} & \textbf{70.5} & \textbf{52.8} & \textbf{26.2} & \textbf{98.6} & \textbf{11.8} & \textbf{78.6} & \textbf{87.6} & \textbf{15.1} & \textbf{43.9} & \textbf{81.6} & \textbf{54.8} \\
    \midrule
    \midrule
    MotionBERT & 2.6 & \textbf{9.1} & \textbf{0.9} & \textbf{3.9} & 7.1 & \textbf{2.0} & \textbf{9.4} & 1.2 & 13.0 & 9.1 & 1.8 & 3.4 & \textbf{8.5} & 5.5 \\
    3D-LFM & 5.3 & 18.5 & 2.3 & 5.7 & 14.1 & 5.2 & 13.8 & 2.3 & 32.3 & 17.0 & 4.8 & 6.6 & 18.9 & 11.6 \\
    Ours & \textbf{2.2} & 10.1 & 1.2 & 4.2 & \textbf{5.0} & 2.5 & 10.8 & \textbf{1.0} & \textbf{12.6} & \textbf{7.8} & \textbf{1.7} & \textbf{3.1} & 8.7 & \textbf{5.4} \\
    \bottomrule
  \end{tabular}
  }}
  \caption{
  \textbf{Quantitative comparison when no artificial noise is applied to the 2D keypoints}. We report, in millimeters, the Sequence-Aligned MPJPE (top), Frame-Aligned MPJPE (middle), and Sequence-Aligned MPVE (bottom). Our approach (Ours) consistently outperforms existing methods across all metrics.
  }
  \label{tab:supp_clean}
\end{table*}

\begin{table}
  \centering
  {\small{
  \begin{tabular}{@{}l|ccc@{}}
    \toprule
    Method & \text{FA-MPJPE} $\downarrow$ & \text{SA-MPJPE} $\downarrow$ & \text{SA-MPVE} $\downarrow$ \\
    \midrule
    MotionBERT & 132.0 & 142.5 & 6.8 \\
    3D-LFM & 112.0 & 143.4 & 21.9 \\
    Ours & \textbf{75.7} & \textbf{86.4} & \textbf{6.4} \\
    \bottomrule
  \end{tabular}
  }}
  \caption{
  \textbf{OOD generalization on unseen data.} We perform a $13$-fold evaluation to assess each method's ability to handle unseen animal categories. Our approach demonstrates superior proficiency in handling OOD 2D-3D lifting.  }
  \label{tab:ood_animal}
\end{table}

\begin{table}
  \centering
  {\small{
  \begin{tabular}{@{}c|ccc@{}}
    \toprule
    Method & FA-MPJPE$\downarrow$ & SA-MPJPE$\downarrow$ & SA-MPVE$\downarrow$ \\
    \midrule
    3D-LFM & 234.4 & 321.7 & 49.0 \\
    Ours & \textbf{143.0} & \textbf{172.9} & \textbf{14.5} \\
    \bottomrule
  \end{tabular}
  }}
  \caption{\textbf{OOD generalization on unseen category and rig.} Models are trained on rigs with 28 or less joints and evaluated on unseen deer and moose categories with 29 joints.
  }
  \label{tab:ood_animal_joints}
\end{table}
\begin{figure}[t]
  \centering
  
  \subfloat[3D-LFM]{
    \includegraphics[width=0.45\linewidth]{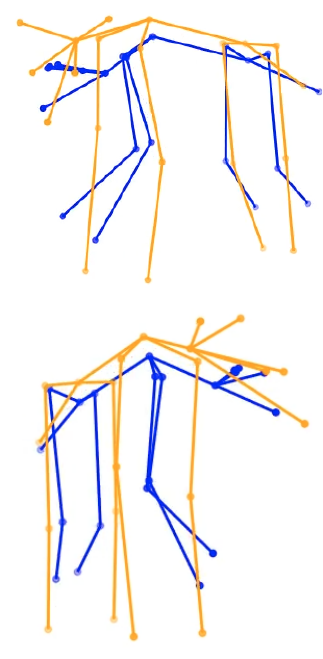}
    \label{fig:3dlfm_bear}
  }
  \hfill
  \vrule
  \hfill
  \subfloat[Ours]{
    \includegraphics[width=0.45\linewidth]{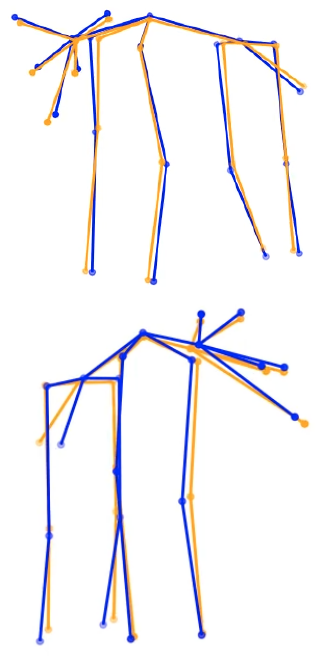}
    \label{fig:ours_bear}
  }

   \caption{\centering
   \textbf{A comparison of robustness in an extreme case of 60\% occlusion.} We showcase the predictions for a bear at 2 different views. Ground truth is orange, prediction is blue.}
   \label{fig:bear}
\end{figure}
\begin{figure}[t]
  \centering
  
  \includegraphics[width=1.0\linewidth]{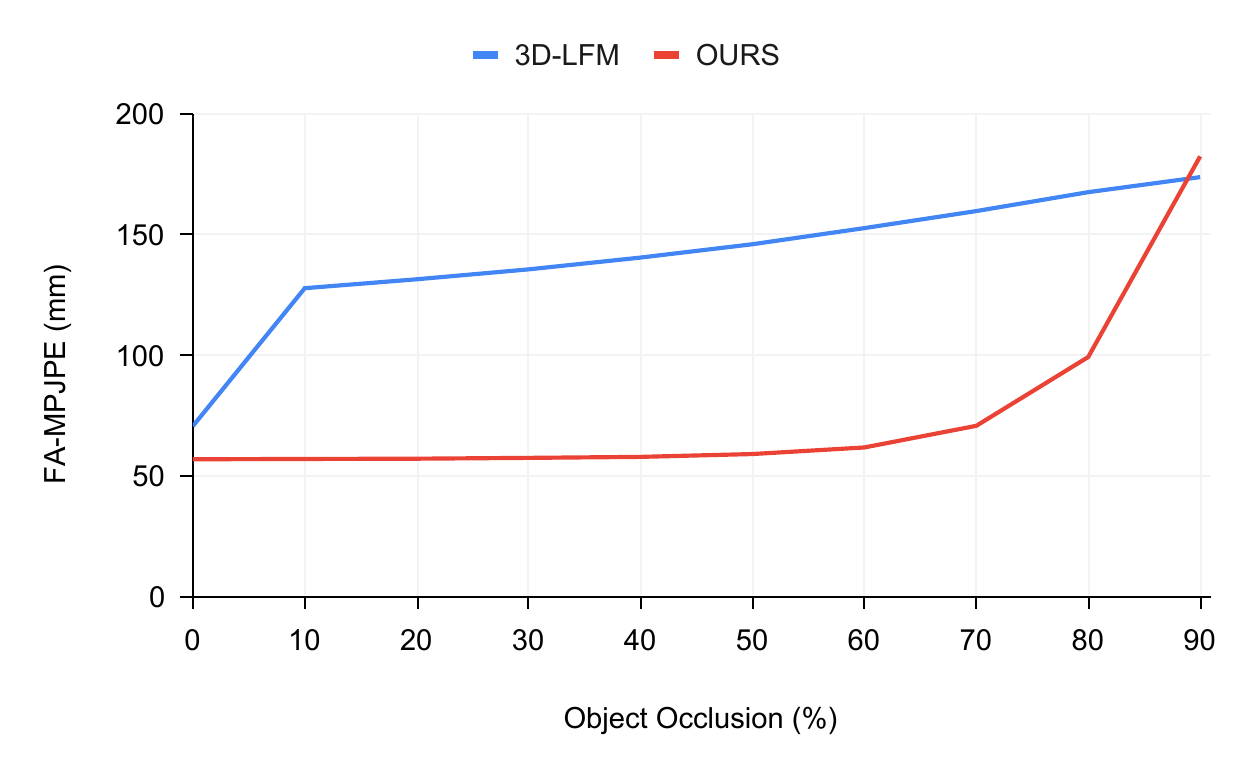}

  \caption{
  \textbf{A comparison of robustness at increasing levels of object occlusion.} Our approach maintains high-fidelity 3D reconstruction, even at extreme levels of occlusion.}
  \label{fig:robustness_chart}
\end{figure}

\begin{table}
  \centering
  {\small{
  \begin{tabular}{@{}l|ccc@{}}
    \toprule
    Method & FA-MPJPE$\downarrow$ & SA-MPJPE$\downarrow$ & SA-MPVE$\downarrow$ \\
    \midrule
    MotionBERT$\dagger$ & 34.8 & 37.2 & 8.8 \\
    MotionBERT & 39.3 & 41.8 & 9.1 \\
    3D-LFM & 48.4 & 63.2 & 29.8 \\
    Ours & 42.9 & 49.0 & 11.1 \\
    \bottomrule
  \end{tabular}
  }}
  \caption{
  \textbf{Results on the Human3.6M benchmark.} MotionBERT$\dagger$ is the original human-specific model. MotionBERT has twice as many parameters and is designed for large-scale human data, while we are designed for small-scale multi-object data.
  }
  \label{tab:supp_human}
\end{table}

\begin{table*}[t]
  \setlength\tabcolsep{4.0pt}
  \centering
  {\small{
  \begin{tabular}{@{}l|c|cccccccccccccc@{}}
    \toprule
    Method & MC & Bear & Buck & Bunny & Chicken & Deer & Dog & Elk & Fox & Moose & Puma & Rabbit & Raccoon & Tiger & Avg \\
    \midrule
    \midrule
    \multirow{2}{*}{MotionBERT} & - & 142.1 & 315.5 & 36.4 & 110.0 & 297.5 & 108.0 & 293.1 & 70.9 & 381.6 & 288.1 & 94.2 & 107.5 & 353.7 & 199.9 \\
    & \checkmark & \textbf{94.5} & \textbf{208.1} & \textbf{16.7} & \textbf{108.2} & \textbf{200.7} & \textbf{50.1} & \textbf{267.4} & \textbf{40.6} & \textbf{189.2} & \textbf{254.4} & \textbf{30.7} & \textbf{77.4} & \textbf{211.8} & \textbf{134.6} \\
    \midrule
    \multirow{2}{*}{3D-LFM} & - & 58.1 & 183.3 & 24.1 & 101.3 & 192.2 & 63.7 & 168.9 & 25.2 & 316.6 & 187.8 & 43.1 & 79.1 & 199.3 & 126.4 \\
    & \checkmark & \textbf{47.6} & \textbf{158.2} & \textbf{23.2} & \textbf{92.3} & \textbf{156.8} & \textbf{53.9} & \textbf{147.8} & \textbf{22.2} & \textbf{274.7} & \textbf{163.4} & \textbf{37.8} & \textbf{70.0} & \textbf{165.4} & \textbf{108.7} \\
    \midrule
    \multirow{2}{*}{Ours} & - & 70.4 & 205.9 & 22.7 & 94.4 & 152.2 & 78.8 & 193.0 & 38.7 & 271.5 & 158.0 & 47.0 & 67.6 & 269.0 & 128.4 \\
    & \checkmark & \textbf{29.2} & \textbf{128.4} & \textbf{17.1} & \textbf{60.8} & \textbf{57.3} & \textbf{32.8} & \textbf{103.1} & \textbf{14.2} & \textbf{97.9} & \textbf{93.2} & \textbf{19.0} & \textbf{44.5} & \textbf{90.8} & \textbf{60.6} \\
    \midrule
    \midrule
    \multirow{2}{*}{MotionBERT} & - & 127.5 & 285.0 & 30.6 & \textbf{96.4} & 264.4 & 90.9 & 263.0 & 63.8 & 305.9 & 258.4 & 93.6 & 98.4 & 309.6 & 176.0 \\
    & \checkmark & \textbf{90.7} & \textbf{198.1} & \textbf{16.0} & 99.0 & \textbf{195.5} & \textbf{45.8} & \textbf{246.8} & \textbf{39.9} & \textbf{170.9} & \textbf{235.0} & \textbf{28.6} & \textbf{74.9} & \textbf{203.2} & \textbf{126.5} \\
    \midrule
    \multirow{2}{*}{3D-LFM} & - & 38.9 & 138.3 & 13.7 & 89.9 & 115.3 & 45.7 & 123.9 & 21.0 & 177.9 & 152.5 & 27.0 & 67.3 & 148.4 & 89.2  \\
    & \checkmark & \textbf{27.9} & \textbf{108.3} & \textbf{12.2} & \textbf{86.3} & \textbf{75.0} & \textbf{33.3} & \textbf{103.0} & \textbf{16.2} & \textbf{119.7} & \textbf{119.3} & \textbf{21.2} & \textbf{57.6} & \textbf{107.7} & \textbf{68.3} \\
    \midrule
    \multirow{2}{*}{Ours} & - & 63.1 & 168.5 & 16.9 & 85.0 & 130.1 & 56.1 & 158.3 & 35.1 & 202.1 & 142.5 & 34.8 & 60.0 & 217.2 & 105.4 \\
    & \checkmark & \textbf{26.7} & \textbf{107.3} & \textbf{11.2} & \textbf{54.2} & \textbf{50.9} & \textbf{27.9} & \textbf{86.1} & \textbf{12.4} & \textbf{81.6} & \textbf{85.9} & \textbf{15.4} & \textbf{42.8} & \textbf{79.8} & \textbf{52.5} \\
    \midrule
    \midrule
    \multirow{2}{*}{MotionBERT} & - & 5.0 & 16.9 & 2.2 & \textbf{3.4} & 12.6 & 5.3 & 12.7 & 2.3 & 26.9 & 12.6 & 4.8 & 5.7 & 16.8 & 9.8 \\
    & \checkmark & \textbf{3.2} & \textbf{11.0} & \textbf{1.1} & 3.9 & \textbf{6.9} & \textbf{2.6} & \textbf{10.4} & \textbf{1.3} & \textbf{17.8} & \textbf{9.8} & \textbf{2.1} & \textbf{4.0} & \textbf{9.8} & \textbf{6.5} \\
    \midrule
    \multirow{2}{*}{3D-LFM} & - & 11.3 & 37.3 & 4.0 & \textbf{3.6} & 30.9 & 10.3 & 29.7 & 4.8 & 51.4 & 29.7 & 7.1 & 13.2 & 32.6 & 20.5 \\
    & \checkmark & \textbf{7.6} & \textbf{29.0} & \textbf{3.4} & 8.4 & \textbf{26.3} & \textbf{8.4} & \textbf{26.5} & \textbf{3.8} & \textbf{43.4} & \textbf{27.3} & \textbf{6.7} & \textbf{12.3} & \textbf{30.4} & \textbf{18.0} \\
    \midrule
    \multirow{2}{*}{Ours} & - & 6.4 & 19.3 & 2.4 & 4.3 & 13.4 & 7.1 & 14.7 & 2.2 & 28.2 & 13.2 & 5.4 & 5.9 & 20.4 & 11.0 \\
    & \checkmark & \textbf{2.5} & \textbf{12.1} & \textbf{1.3} & \textbf{3.4} & \textbf{5.9} & \textbf{2.9} & \textbf{9.3} & \textbf{1.2} & \textbf{12.5} & \textbf{8.9} & \textbf{2.0} & \textbf{3.6} & \textbf{9.1} & \textbf{5.7} \\
    \bottomrule
    \bottomrule
  \end{tabular}
  }}
  \caption{
  \textbf{Per-animal comparison of multi-category and single-category training}. We use a \checkmark for models trained with multiple categories (MC). We report, in millimeters, the Sequence-Aligned MPJPE (top), Frame-Aligned MPJPE (middle), and Sequence-Aligned MPVE (bottom). Top, middle, and bottom are separated by dual horizontal lines.
  }
  \label{tab:supp_multicategory}
\end{table*}
}{} 

\end{document}